\documentclass{ecai} 

\usepackage{latexsym}       
\usepackage{amssymb}        
\usepackage{amsmath}        
\usepackage{amsthm}         
\usepackage{booktabs}       
\usepackage{enumitem}       
\usepackage{graphicx}       
\usepackage{color}          

\usepackage{subfigure}      
\usepackage{pifont}         
\usepackage[table,xcdraw]{xcolor} 
\usepackage{multirow}       
\usepackage{extpfeil}       
\usepackage{xspace}         

\usepackage[colorlinks=true,
            linkcolor=black,     
            citecolor=black,   
            urlcolor=black     
           ]{hyperref}

\usepackage{algorithm}      
\usepackage{algorithmic}    
\usepackage[table]{xcolor}

\newcommand{\cmark}{\checkmark}
\newtheorem{theorem}{Theorem}

\newcommand{\ie}{i.e.}
\newcommand{\eg}{e.g.}

\newcommand{\ourmethod}{Lo-Hp\xspace}
\newcommand{\ourmethodo}{Lo-Op\xspace}

\definecolor{myblue}{HTML}{DDEAF2}

\begin{document}

\begin{frontmatter}


\paperid{7887} 


\title{Learning an Efficient Optimizer via Hybrid-Policy Sub-Trajectory Balance}

\author[A]{\fnms{Yunchuan}~\snm{Guan}}
\author[A\raisebox{-0.8ex}{*}]{\fnms{Yu}~\snm{Liu}}
\author[A]{\fnms{Ke}~\snm{Zhou}}
\author[D]{\fnms{Hui}~\snm{Li}}
\author[I]{\fnms{Sen}~\snm{Jia}} 
\author[B]{\fnms{Zhiqi}~\snm{Shen}} 
\author[F]{\fnms{Ziyang}~\snm{Wang}}
\author[G]{\fnms{Xinglin}~\snm{Zhang}}
\author[H\raisebox{-0.8ex}{*}]{\fnms{Tao}~\snm{Chen}}
\author[E]{\fnms{Jenq-Neng}~\snm{Hwang}}
\author[I]{\fnms{Lei}~\snm{Li}~\thanks{Co-corresponding authors. Emails: liu\_yu@hust.edu.cn, t66chen@uwaterloo.ca, lenny.lilei.cs@gmail.com}}

\begin{abstract}
Recent advances in generative modeling enable neural networks to generate weights without relying on gradient-based optimization. However, current methods are limited by issues of over-coupling and long-horizon. The former tightly binds weight generation with task-specific objectives, thereby limiting the flexibility of the learned optimizer. The latter leads to inefficiency and low accuracy during inference, caused by the lack of local constraints. In this paper, we propose Lo-Hp, a decoupled two-stage weight generation framework that enhances flexibility through learning various optimization policies. It adopts a hybrid-policy sub-trajectory balance objective, which integrates on-policy and off-policy learning to capture local optimization policies. Theoretically, we demonstrate that learning solely local optimization policies can address the long-horizon issue while enhancing the generation of global optimal weights. In addition, we validate Lo-Hp's superior accuracy and inference efficiency in tasks that require frequent weight updates, such as transfer learning, few-shot learning, domain generalization, and large language model adaptation.

\end{abstract}
\end{frontmatter}

\footnotetext[0]{%
\textsuperscript{[A]} Huazhong University of Science and Technology \quad
\textsuperscript{[B]} Nanyang Technological University \quad
\textsuperscript{[C]} National University of Singapore \quad
\textsuperscript{[D]} Jinan Inspur Data Technology Co. \quad
\textsuperscript{[E]} University of Washington \quad
\textsuperscript{[F]} University of Oxford \quad
\textsuperscript{[G]} Shanghai Medical Image Insights Intelligent Technology Co. \quad
\textsuperscript{[H]} University of Waterloo \quad
\textsuperscript{[I]} VitaSight
}



\section{Introduction}\label{sec:intro}
Generative models have rapidly advanced and become central to AI research, driving breakthroughs in areas including vision, audio, language, and structured data~\cite{generative_survey,li2025chatmotion, 10628639, he2025ge}. Recent work has extended the generative model to weight generation for neural networks. They predict downstream neural network weights $\theta$~\cite{he2025enhancing,zhang2023attention,MetaDiff,ma2025energy,jia2024adaptive,liu2025scene,GHN2,GHN3,ji-etal-2024-rag} using a learned forward-only optimizer $f^{G}_{\phi}$, enabling efficient weight adaptation to downstream tasks without computing gradients. Since this innovation can reduce the cost of weight updates, it shows promise in scenarios that require frequent weight updates, such as transfer learning, few-shot learning, domain generalization, and LLM fine-tuning. However, existing methods are limited by two issues: over-coupling and long horizon.

\textbf{Over-Coupling:} Previous studies, such as Meta-HyperNetwork~\cite{Meta-Hypernetwork} and the GHN series~\cite{GHN,GHN2,GHN3,11144414}, adopted an end-to-end approach that directly models downstream task performance. These methods define the optimization objective as
\begin{equation}\label{eq:end-to-end optimization}
     \arg\min_{\phi} \underset{(x,y)\in \mathcal{D}}{E} \bigl[ L_D(f_{\theta}(x;\theta=f^G_{\phi}(x;\phi)),y)\bigr],
\end{equation}
where $L_D$ refers to the loss function of downstream tasks and $\mathcal{D}$ denotes the dataset. Although straightforward, this approach suffers from over-coupling. It binds the objectives of weight generation and downstream task optimization, thus limiting the flexibility for the learned optimizer $f^G_{\phi}$.\footnote{The terms generative model and learned optimizer are used interchangeably.} Specifically, the inference process of $f^G_{\phi}$ must be differentiable and can only be unrolled over a short horizon. This prevents the learned $f^G_{\phi}$ from capturing more expressive policies over long horizons.

\begin{figure}[t]
    \centering
    \includegraphics[width=0.6\linewidth]{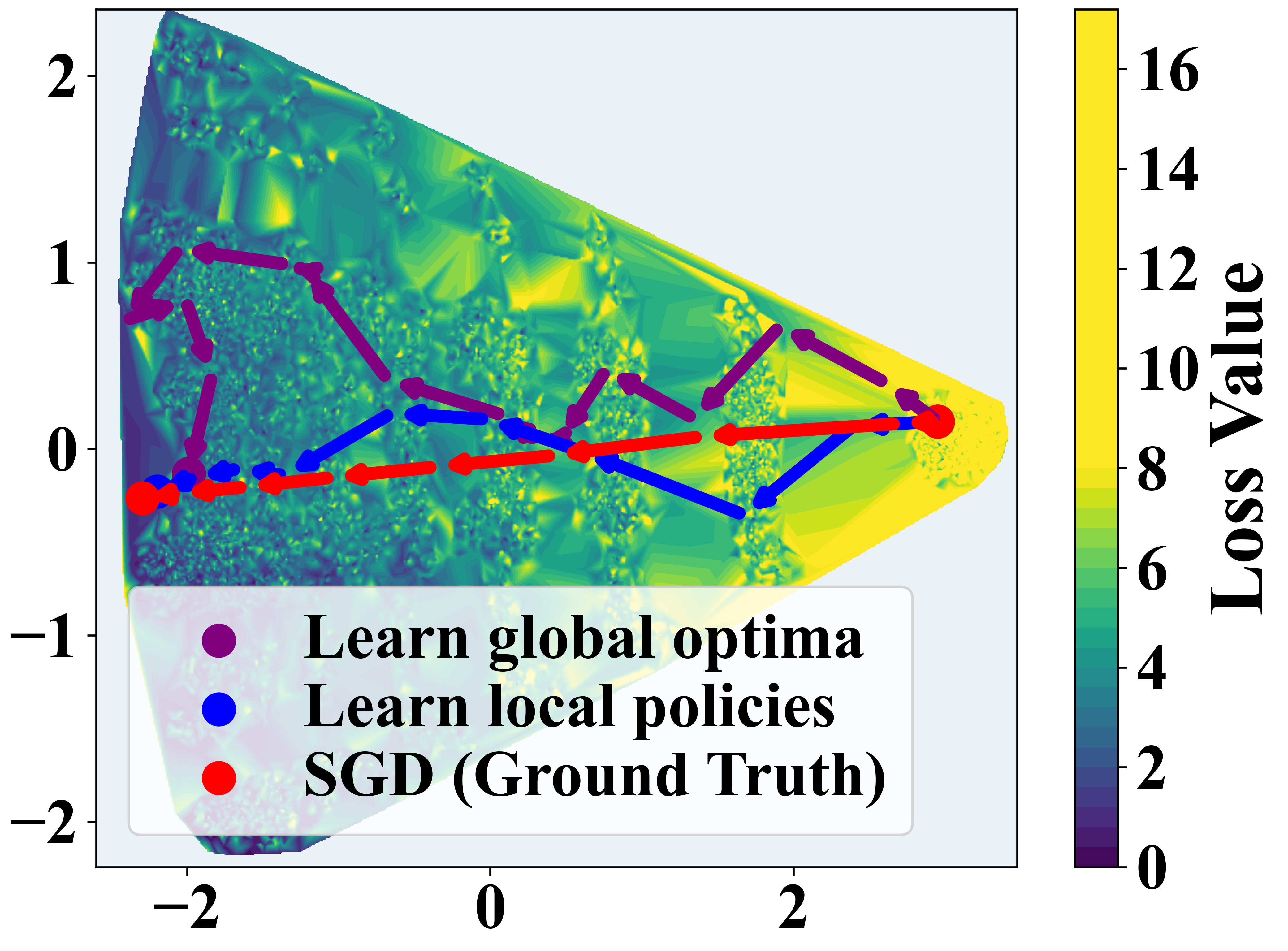}
    \caption{Inference trajectory of generative models in CIFAR-10's 2D weight-reduced space. Darker regions indicate lower downstream task loss, and the red trajectory represents the ground truth generated by the real-world optimizer SGD.}
    \vspace{2em}
    \label{fig:landscape_2d}
\end{figure}

\begin{figure*}
    \centering
    \includegraphics[width=0.6\linewidth]{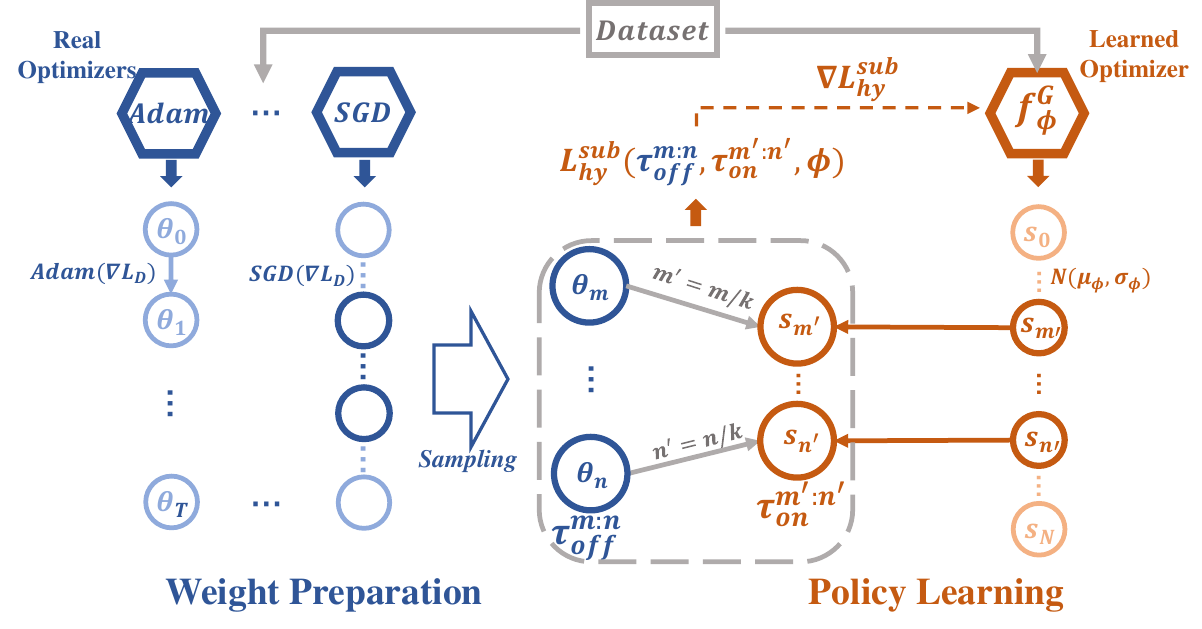}
    \caption{Overview of \ourmethod. It consists of two decoupled stages: weight preparation and policy learning. In the weight preparation stage, it utilizes learned optimizers such as Adam, SGD, etc., to update the neural network weights $\theta$. Then, it samples and records the offline sub-trajectory $\tau^{m:n}_{off}$. In the policy learning stage, the generative model $f^{G}_{\phi}$ adopts a Gaussian policy to generate the online trajectory. A uniform sub-trajectory matching strategy is used to align the online sub-trajectory $\tau^{m':n'}_{on}$ and offline sub-trajectory $\tau^{m:n}_{off}$, and the proposed hybrid-policy sub-trajectory balance is applied to learn local optimization policies.}
    \vspace{1em}
    \label{fig:overview}
\end{figure*}

\textbf{Long-Horizon:} Recent studies, such as OCD~\cite{OCD}, MetaDiff~\cite{MetaDiff}, and D2NWG~\cite{D2NWG}, leverage models such as diffusion to model the downstream neural network weights $\theta$, enabling more faithful simulation of optimizer behaviors over a long-horizon inference. However, these methods focus on a single optimization policy (\eg, SGD), thereby failing to explore the flexibility of this paradigm. More importantly, they focus solely on the global optima $\theta_*$, neglecting the local policy details in the sub-trajectories $\theta_m \to \theta_n$. As shown by the purple trajectory in Figure~\ref{fig:landscape_2d}, such an unconstrained learning process leads to low efficiency and low accuracy in the inference trajectories over a long horizon.

In this paper, we rethink weight generation as an optimization policy learning problem. Our objective is to develop methodologies for generating global optimal weights as well as modeling local optimization policies. We propose to \textbf{L}earn an efficient \textbf{O}ptizer via \textbf{H}ybrid-\textbf{P}olicy Sub-Trajectory Balance, \ie, Lo-Hp. (1) To address the limited flexibility caused by over-coupling, \ourmethod adopts a decoupled two-stage learning process that enables the learning of diverse local optimization policies. Formally, the decoupled learning framework can be defined as
\begin{align}\label{eq:decoupled objective}
    \{\theta_t\}_0^T=\arg\min_{\theta} \underset{{(x,y)\in \mathcal{D}}}{E} \bigl[L_D (x,y,f_\theta)\bigr]\notag\\
    \phi=\arg\min_{\phi}\underset{x,m,n}{E}\bigl[\mathcal{L}^{sub}_{hy}(x,\{\theta_t\}_m^n,f^G_{\phi})\bigr].
\end{align}

As shown in Figure~\ref{fig:overview}, the weight preparation stage uses multiple optimizers (\eg, SGD and Adam) to construct diverse offline trajectories, enhancing the policy flexibility of the learned $f^G_{\phi}$. In the policy learning stage, the generative model $f^G_{\phi}$ adopts Hybrid-Policy Sub-Trajectory Balance to constrain the inference trajectory at the local level. (2) Hybrid-Policy Sub-Trajectory Balance is designed to address the inefficiency in inference caused by the long horizon. It is a hybrid learning strategy that lies between on-policy and off-policy learning~\cite{on-off-policy, zhou2025reagent,yao2025countllm}. It introduces supervision signals from the offline sub-trajectory, \ie, $\{\theta_t\}_m^n$, into the learning process of the online sub-trajectory, \ie, $\{s_i\}_{m'}^{n'}$,\footnote{The terms online trajectory, inference trajectory, and sampling trajectory are used interchangeably.} thereby enabling $f^G_{\phi}$ to acquire optimization policies at the local level. As shown by the blue trajectory in Figure~\ref{fig:landscape_2d}, this approach improves both the efficiency and accuracy of the weight generation process. (3) We theoretically show that our method, though focusing solely on local optimization policy learning, remains capable of promoting the generation of global optimal weights. In addition, we analyze the convergence of our method and introduce Sharpness Aware Minimization (SAM)~\cite{SAM}. It improves the overall convergence efficiency of the framework. Extensive experiments demonstrate Lo-Hp's superior accuracy and inference efficiency in tasks that require frequent weight updates. Our contributions can be summarized as follows:
\vspace{-3pt}
\begin{itemize}
    \item We propose a decoupled framework \ourmethod, which consists of weight preparation and policy learning stages. It enhances the flexibility of the learned optimizer by learning various optimization policies.
    \item We propose a hybrid-policy sub-trajectory balance, which captures local policy details while simultaneously facilitating the generation of globally optimal weights.
    \item We analyze the convergence of the decoupled weight generation framework and introduce SAM to enhance convergence efficiency.
\end{itemize}

\section{Method}
The proposed \ourmethod decouples the learning process into two stages: weight preparation and policy learning. This framework enhances the flexibility of the learned optimizer. During the weight preparation stage, \ourmethod constructs offline trajectories using various optimization policies. During the policy learning stage, the generative model $f^G_{\phi}$ adopts Hybrid-Policy Sub-Trajectory Balance to capture local optimization policies.
\subsection{Weight Preparation}
As shown on the left side of Figure~\ref{fig:overview}, the target of the weight preparation stage is to collect optimization trajectories constructed by different optimization policies. Formally, this process can be defined as
\begin{equation}\label{eq:weighet_preparation}
    \tau^{0:T}_{of}=\{\theta_t\}_0^T=\arg\min_{\theta} \underset{{(x,y)\in \mathcal{D}}}{E} \bigl[L_D (x,y,f_\theta)\bigr].
\end{equation}

Since these trajectories are built on real-world optimizers, we refer to them as offline trajectories, \ie, $\tau_{of}^{0:T}=\{\theta_t\}_0^T$. Specifically, multiple optimizers are used to solve Equation~\ref{eq:weighet_preparation}, from which we collect the checkpoint weights $\theta_t$ and the associated data $x$. Within an offline trajectory, $\theta_0$ denotes the Gaussian-initialized weight, $\theta_T$ represents the global optimum, and $T$ is the total number of training epochs. In our implementation, we use two optimizers, \ie, Adam and SGD, and an auxiliary optimization policy, \ie, Sharpness Aware Minimization (SAM). Specifically, SGD is well-suited for large, clean datasets such as ImageNet, while Adam excels in fast convergence tasks like few-shot learning on Mini-ImageNet. The motivation and role of SAM are detailed in Section~\ref{sec:SAM}.

\begin{algorithm}[t]
\caption{Weight Preparation}
\label{alg:SAM}
\begin{algorithmic}[1]
    \REQUIRE Downstream task weights $\theta_0$, downstream task loss $L_D$, perturbation $\rho$, optimization policy $Op$, dataset $\mathcal{D}$.
    \FOR{t in range(T)}
        \STATE Sample batch of data $S_i \sim \mathcal{D}$ 
        \STATE $\epsilon \gets \rho \frac{\nabla L_D(S_i;\theta_t)}{\|\nabla L_D(S_i;\theta_t)\|}$
        \STATE $g_{SAM} \gets \nabla_{\theta} L_D(S_i;\theta_t + \epsilon)$
        \STATE $\theta_{t+1} = Op(\theta_t, g_{SAM})$
        \STATE Append $\theta_t$ to $\tau_{off}$
    \ENDFOR
    \STATE Randomly sample sub-trajectories $\tau^{m:n}_{of}$ from $\tau_{off}$
    \RETURN $\{\tau^{m:n}_{off}\}_{m \in [0,T), n \in (m,T]}$
\end{algorithmic}
\end{algorithm}

Compared to the end-to-end optimization objective given by Equation~\ref{eq:end-to-end optimization}, the decoupled weight preparation offers more choices for the optimization policy. By improving the flexibility of optimization policies, it further enhances the robustness of the policy learned by the generative model $f_{\phi}^G$. Our experiments in Section~\ref{sec:vairous policy experiment} support the above claim. The concern about overhead is detailed in the Supplementary Material C.3. The specific process of weight preparation is detailed in Algorithm~\ref{alg:SAM}.

\subsection{Policy Learning}
As illustrated on the right side of Figure~\ref{fig:overview}, we adopt a learnable Gaussian policy for inference in continuous weight space. Formally, the online inference trajectory, \ie, $\tau_{on}^{0:N}=\{s_i\}_0^N$, starts from $s_0=\theta_0 \sim N(0,1)$ and is driven by
\begin{equation}\label{eq:gaussian_policy}
    s_{t+1} \sim \mathcal{N}(\mu_\phi(s_t), \sigma_\phi(s_t)).
\end{equation}
It can be found that the inference trajectory is determined solely by the Gaussian policy, ignoring the policy details behind real-world offline trajectories. We refer to this on-policy method, which models only the global optima, as \ourmethodo. As shown by the purple trajectory in Figure~\ref{fig:landscape_2d}, \ourmethodo's unconstrained inference trajectory exhibits low efficiency and poor accuracy over a long horizon. Previous methods, such as OCD, MetaDiff, and D2NWG, fall into this category. In this paper, we propose Hybrid-Policy Sub-Trajectory Balance to introduce a supervision signal from offline sub-trajectories to the learning process of online sub-trajectories. It captures local optimization policies and enables better efficiency and accuracy. For the generative model $f^G_{\phi}$, we use the commonly used U-Net architecture, conditioning on the unlabeled samples $\{x_i\} \in \mathcal{D}$ to differentiate among tasks (Equation~\ref{eq:decoupled objective}). The specific process of policy learning is detailed in Algorithm~\ref{alg:policy training}.

\begin{algorithm}[t]
\caption{Policy Learning}
\label{alg:policy training}
\begin{algorithmic}[1]
    \REQUIRE Generative model $f^{G}_\phi$, learning rate $\alpha$, sub-trajectories $\{\tau^{m:n}_{off}\}_{m \in [0,T), n \in (m,T]}$
    \WHILE{not converged}
        \STATE Select offline sub-trajectory $\tau^{m:n}_{off}$
        \STATE Start from $s_0=\theta_0$, and sample online trajectory $\tau^{0:N}_{on}$  (Equation~\ref{eq:gaussian_policy})
        \STATE Match $\tau^{m':n'}_{on}$ for $\tau^{m:n}_{off}$ (Equation~\ref{eq:trajectory_match})
        \STATE Compute $\nabla_{\phi}\mathcal{L}^{sub}_{hy}(\tau^{m':n'}_{on},\tau^{m:n}_{off};\phi)$
        \STATE Update $\phi \gets \phi - \alpha \nabla \mathcal{L}^{sub}_{hy}$
    \ENDWHILE
\end{algorithmic}
\end{algorithm}

\subsubsection{Hybrid-Policy Sub-Trajectory Balance}
For an online sub-trajectory $\tau_{on}^{m':n'}=\{s_t\}_{m'}^{n'}$, the loss function of vanilla sub-trajectory balance~\cite{TB}(sub-TB) can be written as
\begin{align}\label{eq:vanilla_sub-TB}
\mathcal{L}^{sub} = \underset{m',n'}{E}~\Bigl\| &\log F_{\phi}(s_{m'}) + \sum_{t=m'}^{n'-1} \log P^F_{\phi}(s_{t+1} \mid s_t) \notag\\
- &\log F_{\phi}(s_n') - \sum_{t=m'}^{n'-1} \log P^B_{\phi}(s_t \mid s_{t+1}) \Bigl\|_2^2,
\end{align}
where the generative model is defined as $f^G_{\phi}:=\{P^F_{\phi},P^B_{\phi},F_{\phi}\}$, $P^F_{\phi}$ is the learnable forward Gaussian policy, $P^B_{\phi}$ is the learnable backward Gaussian policy, and $F_{\phi}$ is the flow function. Following sub-TB, $P^F_{\phi}$, $P^B_{\phi}$, and $F_{\phi}$ are parameterized by $\phi$, utilizing a shared bottleneck for representation learning and separate heads to distinguish the forward policy, backward policy, and flow function, respectively.\footnote{For simplicity, we use the same symbol $\phi$ to represent the parameters of $P^F$, $P^B$, and $F$, disregarding the differences in their heads.} The conditional probability $P_{\phi}(s_{t+1} \mid s_t)$ is given by
\resizebox{\linewidth}{!}{%
\begin{minipage}{\linewidth}
\begin{align}
\log P_{\phi}(s_{t+1} \mid s_t) 
= & -\frac{1}{2}(s_{t+1} - \mu_{\phi}(s_t))^\top \sigma_{\phi}^{-1}(s_t) (s_{t+1} - \mu_{\phi}(s_t)) \notag \\
  & - \frac{1}{2} \log\left[(2\pi)^d \det(\sigma_{\phi}(s_t))\right].
\end{align}
\vspace{1pt}
\end{minipage}
}

Given an offline sub-trajectory $\tau_{off}^{m:n}=\{\theta_t\}_{m}^n$, we define our hybrid-policy loss as \footnote{It is worth noting that we adopt \(\|\cdot\|_2\) instead of \(\|\cdot\|_2^2\) to enable the use of the triangle inequality in the proof of Theorem~\ref{thm:full-tra}, thereby guaranteeing the desired global property.
}\\
\noindent\resizebox{\linewidth}{!}{%
\begin{minipage}{\linewidth}
\begin{align}
\mathcal{L}^{sub}_{hy} = \underset{m',n'}{E}~\Bigl\| &\log C_{\phi}(s_{m'})R_n(s_{m'}) + \sum_{t=m'}^{n'-1} \log P^F_{\phi}(s_{t+1} \mid s_t) \notag\\
-&\log C_{\phi}(s_{n'})R_n(s_{n'}) - \sum_{t=m'}^{n'-1} \log P^B_{\phi}(s_t \mid s_{t+1}) \Bigl\|_2,
\label{eq:our_loss}
\end{align}
\end{minipage}
}
where $C_{\phi}(\cdot)$ is a learnable coefficient and $R_n$ is the reward function
\begin{equation}
    R_n(s_t)=e^{-||s_t-\theta_n||_2^2}.
\end{equation}
We replace the flow function $F_{\phi}(s_t)$ in vanilla sub-TB with $C_{\phi}(s_t)R_n(s_t)$, since the reward for intermediate states can be directly evaluated in our setting (\ie, weight generation). $R_n(s_t)$ describes how close the current state is to the local target $\theta_n$. The matching strategy for the local target is detailed in Section~\ref{sec:matching}. 

Moreover, according to the principle of TB, the replaced $F(s_t)$ is proportional to $R_n(s_t)$. To ensure consistency with this principle, we introduce a learnable coefficient $C_{\phi}(\cdot)$.\footnote{Similar to $P^F$ and $P^B$, we use the same symbol $\phi$ to denote the parameters, and omit the differences between the heads.} The theoretical soundness of the above design is established through the following theorems.

\begin{theorem}\label{thm:sub-tra}
Suppose that $\mathcal{L}_{hy}^{sub} = 0$.
Then, the expected cumulative probability of the sub-inference trajectories \(\tau^{m':n'}_{on}=\{s_{m'},\cdots, s_{n'}\}\) satisfies
\[
\underset{{\tau\in\mathcal{T}_{m':n'}}}{E}\prod_{t=m'}^{n'-1}P^F_{\phi}(s_{t+1}\mid s_t) \;\propto\; R_{n}(s_{n'}).
\]
\end{theorem}
\begin{theorem}\label{thm:full-tra}
Suppose that $\mathcal{L}_{hy}^{sub} = 0$.
Then, the expected cumulative probability of the full inference trajectory $\tau^{0:N}_{on}=\{s_0,...,s_N\}$ satisfies
\[
\underset{{\tau\in\mathcal{T}_{0:N}}}{E}\prod_{t=1}^{N-1} P^F_{\phi}(s_{t+1}\mid s_{t}) \;\propto\; R_{T}(s_N).
\]
\end{theorem}

The proof is detailed in the Supplementary Material. The above theorems indicate that our proposed objective $\mathcal{L}^{sub}_{hy}$ possesses the following properties:
\begin{enumerate}
    \item For sub-trajectories, their cumulative probability is proportional to \(R_n(s_{n'})\). This indicates that the inferred sub-trajectory's endpoint is very close to the local target $\theta_n$, suggesting that the learned optimizer $f_{\phi}^G$ captures the local optimization policy.
    \item For full trajectories, their cumulative probability is proportional to \(R_T(s_N)\). This indicates that the inferred full trajectory's endpoint is very close to the global optimum $\theta_T$, suggesting that the learned optimizer enables the generation of optimal weights.
\end{enumerate}
Therefore, \ourmethod can learn a local policy to guide sampling trajectory while facilitating the generation of global optimal weights.

\subsubsection{Trajectory Matching}\label{sec:matching}
As shown in Figure~\ref{fig:overview}, the lengths of the online trajectory $T$ and offline trajectory $N$ are inconsistent. To achieve better sub-trajectory matching, we adopt a uniform assignment strategy in our implementation. More precisely, we enforce the length of the online trajectory to satisfy $T = kN$, where $k$ is a segmentation factor. Formally, any offline sub-trajectory $Tra_{off}^{sub}$ and its corresponding online sub-trajectory $Tra_{on}^{sub}$ satisfy the following relation:
\begin{equation}\label{eq:trajectory_match}
    \tau_{off}^{m:n}=\{ \theta_m, \ldots, \theta_n \} \xleftrightarrow{\text{match}} \tau_{on}^{m':n'}=\{ s_{m'=\frac{m}{k}}, \ldots, s_{n'=\frac{n}{k}} \}.
\end{equation}

In addition to gradient-free optimization, this design allows~\ourmethod to solve in the weight space $k$ times faster than real-world optimizers. As a result, during the weight preparation stage, we can use a very small learning rate to ensure the stability and accuracy of offline trajectories. On the other hand, we can avoid the overhead caused by overly long online trajectories.

\section{Discussion}
In this section, we discuss how to improve \ourmethod's efficiency by $k$. We also analyze Lo-Hp's improvement on the local optimization policy. Furthermore, we provide a convergence analysis of such a decoupled weight generation framework and introduce SAM to improve convergence efficiency.

\subsection{Efficiency Improvement}
According to the matching strategy provided by Equation~\ref{eq:trajectory_match}, we can increase $k$ to reduce the sampling trajectory length $N$, thereby further improving inference efficiency. Table~\ref{tab:k_ablation} shows how the accuracy and inference latency of \ourmethod vary with different values of $k$. We evaluate on the CIFAR-10 transfer learning task and the Mini-ImageNet 5-way 1-shot task (see Section~\ref{sec:Comparison Experiments} for setup details). As $k$ increases, inference latency decreases due to fewer online states $s_t$ being used to estimate the offline sub-trajectories (Equation~\ref{eq:trajectory_match}). However, this comes at the cost of reduced accuracy. To balance accuracy and latency, we set $k=2$ in the following experiments, achieving a $2\times$ speedup.

\begin{table}[t]
\label{tab:k_ablation}
\caption{Accuracy and inference latency under different values of $k$ on CIFAR-10 transfer learning task and Mini-ImageNet 5-way 1-shot task.}
\vspace{1.5em}
\centering
\begin{tabular}{cccc}
\toprule
 & \textbf{CIFAR-10} & \textbf{Mini-ImageNet} & \textbf{Latency(ms)} \\
\midrule
$k=1/2$ & 65.11 ± 0.25 & 66.19 ± 0.27 & 12.6\\
$k=1$ & 64.97 ± 0.22 & 66.04 ± 0.29 & 6.1\\
\rowcolor{myblue}$k=2$ & 64.25 ± 0.28 & 65.21 ± 0.36 & 2.9\\
$k=3$ & 62.58 ± 0.56 & 63.87 ± 0.60 & 2.1\\
\bottomrule
\end{tabular}
\end{table}%

\begin{figure}[t]
\centering
\resizebox{\linewidth}{!}{\includegraphics{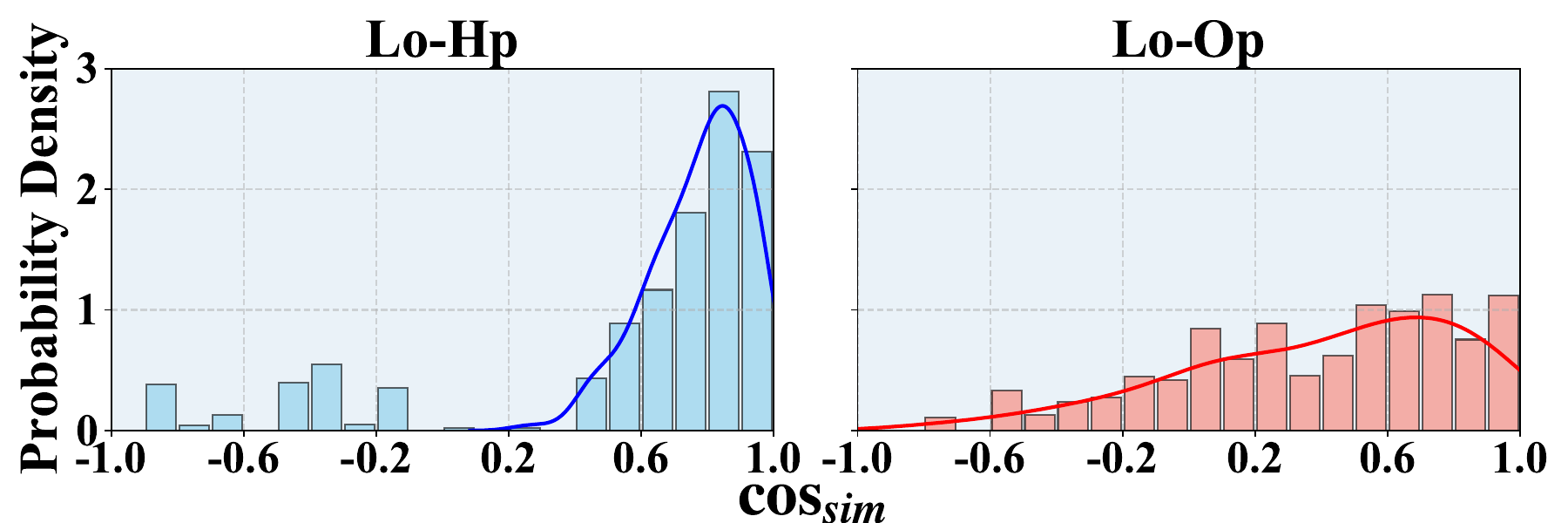}}
\vspace{-1em}
\caption{Similarity statistics between generated online sub-trajectories and target offline sub-trajectories on CIFAR-10.}
\vspace{2em}
\label{fig:local_policy_details}
\end{figure}

\subsection{Local Policy Details}
The proposed \ourmethod aims to capture local policy details through hybrid policy sub-trajectory learning. Therefore, beyond the final optimal weight, we are also interested in whether \ourmethod exhibits optimizer-like behavior at the local level. To measure whether a learned optimizer $f^G_{\phi}$ behaves like a real-world optimizer, we compute the cosine similarity between its generated online sub-trajectories $\tau_{on}^{m' \to n'}$ and the offline target $\tau_{off}^{m \to n}$ as
\begin{equation}
    \cos_{sim}=\frac{(s_{n'}-s_{m'})(\theta_{n}-\theta_{m})}{||s_{n'}-s_{m'}||{\cdot||\theta_{n}-\theta_{m}||}}.
\end{equation}

Figure~\ref{fig:local_policy_details} shows the distribution of \ourmethod, which learns local policy details, and \ourmethodo, which models only global optima. \ourmethod yields a more concentrated distribution near 1, while \ourmethodo is more dispersed, indicating that \ourmethod aligns better with real optimizers even at the local level.

\subsection{Convergence Analysis and Improvement}\label{sec:SAM}
Unlike previous end-to-end frameworks, the decoupled framework used here involves multiple independent losses and models, making convergence harder to guarantee. We next derive its empirical error bound and introduce improvements.

\begin{theorem}\label{theorem:emperi error}
When the reconstruction error of the generative model is bounded by \( c \), the downstream loss satisfies \( L_d(\cdot) \leq \psi \), and the loss function is both \( l \)-smooth and \( \mu \)-strongly convex, with the eigenvalues of the Hessian matrix around the optimum \( \theta_* \) bounded by \( \lambda \), the cumulative empirical error of the decoupled weight generation framework can be bounded as follows
\begin{equation}
    L_D(\hat{\theta})-L_D(\theta^*) \leq \frac{\lambda}{2} \left[c+\frac{2\psi}{\mu}\left(1 - \frac{\mu}{l}\right)^{T}\right],
\end{equation}
where $\hat{\theta}$ is the weight predicted by the generative model.
\end{theorem}
The proof, provided in the Supplementary Material, relies on the triangle inequality to decompose the accumulated error into weight preparation error and reconstruction error. We make a $\mu$-strong convex assumption here, but subsequent analysis and improvement do not rely on this property, thus preserving the practicality of our derivation. Theorem~\ref{theorem:emperi error} shows that, compared to direct learning methods, the reconstruction error of weight generation algorithms affects the upper bound of cumulative error in only a linear manner. Furthermore, this upper bound can be effectively improved by reducing the maximum eigenvalue $\lambda$. Penalizing the Hessian matrix is the simplest way to accelerate convergence, but it is computationally unacceptable. 

In this paper, we penalize $\lambda$ by constraining the curvature near the neighborhood of the optimal solution. We use Sharpness-Aware Minimization (SAM)~\cite{SAM} in the weight preparation stage to achieve the above target. The process of SAM is shown in Algorithm~\ref{alg:SAM}.

\section{Experiment}
Our experimental platform includes two A100 GPUs, one Intel Xeon Gold 6348 processor, and 512 GB of DDR4 memory. For all experiment results, we report the mean and standard deviation over 5 independent repeated experiments. We present the basic experimental results, with more setup details provided in the Supplementary Material.

\subsection{Ablation Study}
\subsubsection{Main Components}\label{sec:ablation_main}

The advantages of \ourmethod stem from three main components:
\begin{itemize}[align=left]
    \item C1: Using a decoupled weight generation framework to learn a more flexible optimization policy.
    \item C2: Using the trajectory balance loss to model the global optimal weights.
    \item C3: Using the proposed hybrid policy sub-trajectory balance loss to introduce offline supervision signals, thereby enabling the learning of local optimization policies.
\end{itemize}

Note that C2 builds upon C1, and C3 builds upon C2. Therefore, we conduct incremental ablation experiments here. As shown in Table~\ref{tab:ablation_main}, we validate the effectiveness of each component on CIFAR-10 transfer learning and Mini-ImageNet 5-way 1-shot tasks. We use four convolution blocks with a linear probe for classification. When none of the components are used, our method degrades to a Hypernetwork optimized by an end-to-end objective, \ie, Equation~\ref{eq:end-to-end optimization}. When only C1 is used, we employ a commonly used diffusion algorithm (rather than trajectory balance) to model the global optimal weights, which we refer to as Lo-Di. When only C1 and C2 are used and the trajectory balance is applied to model global optimal weights, our method reduces to \ourmethodo.

\begin{table}[t]
\centering
\caption{Ablation main components on CIFAR-10 transfer learning task and Mini-Imagenet 5-way 1-shot task. Metric by accuracy.}
\vspace{1.5em}
\resizebox{\linewidth}{!}{
\begin{tabular}{lcccccc}
    \toprule
    & \textbf{C1} & \textbf{C2} & \textbf{C3} & \textbf{CIFAR-10} & \textbf{Mini-Imagenet} \\
    \midrule
    Hypernetwork    &             &             &             & 43.71 ± 0.20 & 45.29 ± 0.28 \\
    Lo-Di    & \cmark      &             &             & 61.67 ± 0.14 & 61.84 ± 0.30 \\
    \ourmethodo       & \cmark      & \cmark      &             & 61.28 ± 0.32 & 62.53 ± 0.37 \\
    \ourmethod     & \cmark      & \cmark      & \cmark      & 64.25 ± 0.28 & 65.21 ± 0.36 \\
    \bottomrule
\end{tabular}}
\label{tab:ablation_main}
\end{table}

\begin{figure}[t]
    \centering
        \includegraphics[width=\linewidth]{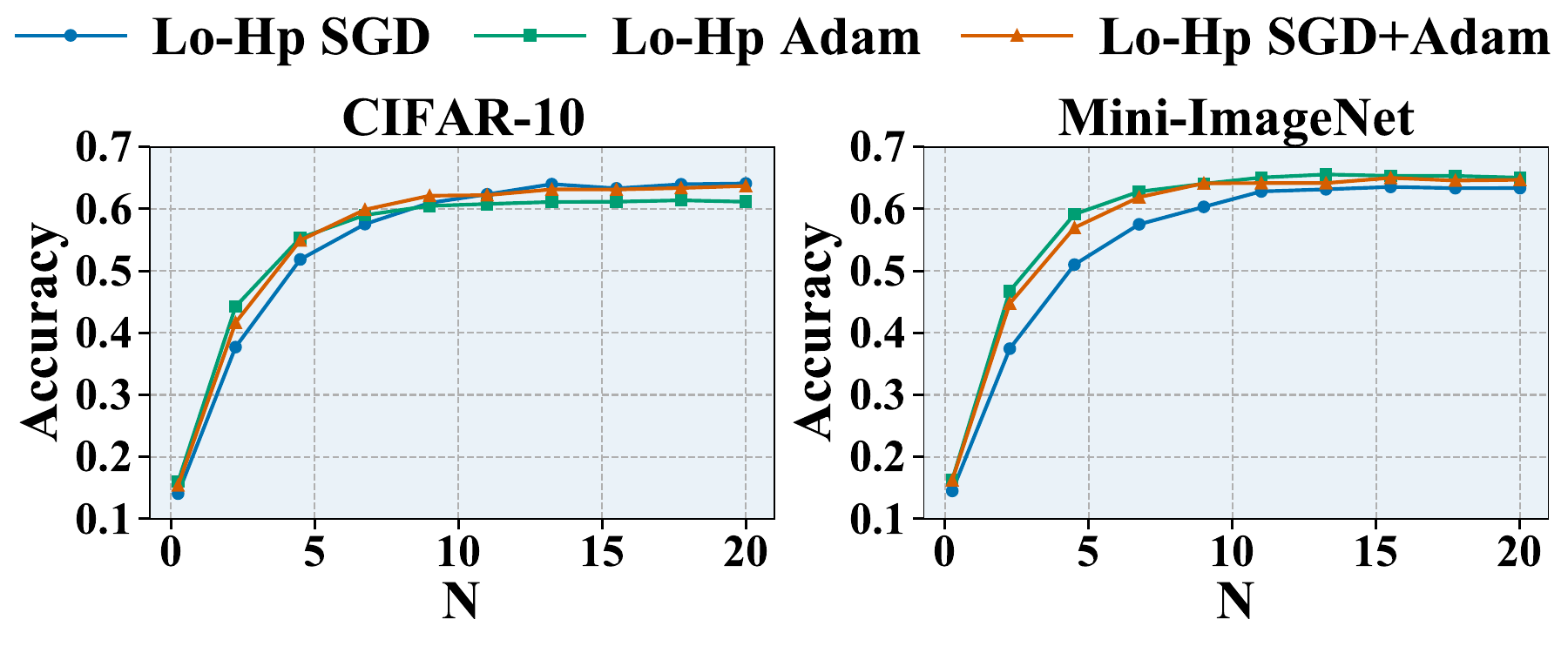}
        \vspace{-2em}
        \caption{The impact of different offline optimization policies on \ourmethod's inference curve.}
        \vspace{2em}
        \label{fig:policy_ablation}
\end{figure}

\begin{figure}[t]
    \centering
    \includegraphics[width=\linewidth]{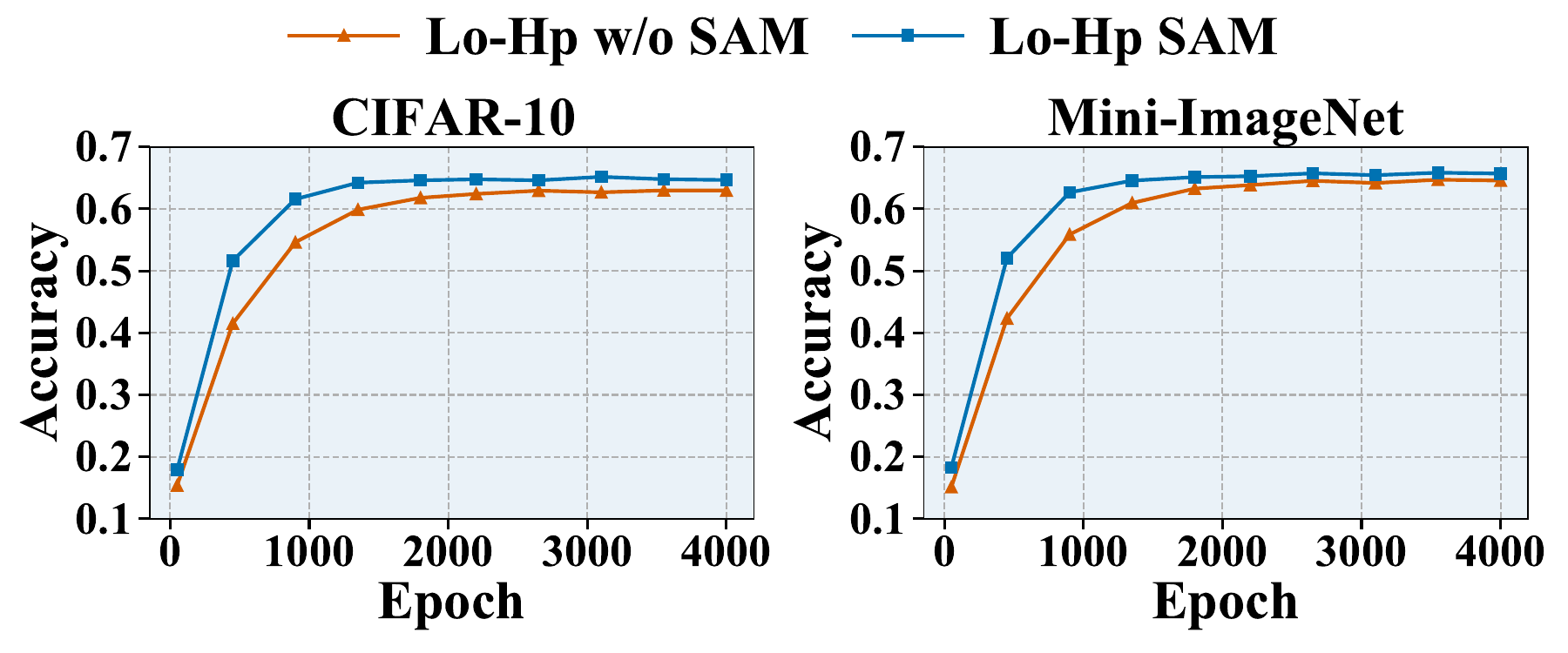}
    \vspace{-1.5em}
    \caption{The impact of SAM on \ourmethod's learning curve.}
    \vspace{2em}
    \label{fig:SAM_ablation}
\end{figure}

As shown in Table~\ref{tab:ablation_main}, Lo-Di and \ourmethodo demonstrate higher accuracy compared to Hypernetwork, suggesting the necessity for a decoupled framework. The comparison between Lo-Di and \ourmethodo shows that using different generative models does not significantly affect the accuracy on downstream tasks. However, once hybrid-policy sub-trajectory learning is incorporated, \ourmethod surpasses both \ourmethodo and Lo-Di, which can be attributed to its capability of learning local optimization policies.

\subsubsection{Offline Optimization Policies}\label{sec:vairous policy experiment}
In the weight preparation stage, we use two optimizers, \ie, SGD and Adam, to improve policy flexibility. Specifically, SGD is well-suited for large, clean datasets such as ImageNet, while Adam excels in fast-convergence tasks like few-shot learning on Mini-ImageNet. Figure~\ref{fig:policy_ablation} shows the learned optimizer's inference curves produced by three weight preparation schemes, \ie, SGD, Adam, and SGD+Adam, on CIFAR-10 and Mini-ImageNet. Although not consistently superior in all periods, the offline optimization trajectories that the combination of SGD and Adam enables \ourmethod to achieve a more balanced inference speed and accuracy across diverse tasks.

To improve convergence efficiency, we introduce SAM. Figure~\ref{fig:SAM_ablation} shows the learning curves of two schemes, \ie, \ourmethod w/o SAM and standard \ourmethod. It can be observed that SAM not only improves the overall convergence efficiency of \ourmethod during training but also enhances downstream task accuracy on the CIFAR-10 transfer learning task.

\subsection{Comparison Experiments}\label{sec:Comparison Experiments}

\begin{table*}[t]
\caption{Transfer learning accuracy comparison on various datasets with average per-sample evaluation latency.}
\vspace{1.5em}
\centering
\resizebox{\linewidth}{!}{
\begin{tabular}{llllllll}
\hline
\textbf{Learning Type} & \textbf{Method} & \textbf{CIFAR-10} & \textbf{CIFAR-100} & \textbf{STL-10} & \textbf{Aircraft} & \textbf{Pets} & \textbf{Latency (ms)} \\ \hline
Gradient-Based & ICIS~\cite{ICIS} & 61.75 ± 0.31 & 47.66 ± 0.24 & 80.59 ± 0.12 & 26.42 ± 1.56 & 28.71 ± 1.60 & 9.2 \\
End-to-End & GHN3~\cite{GHN3} & 51.80 ± 0.42 & 11.90 ± 0.45 & 75.37 ± 0.19 & 23.19 ± 1.38 & 27.16 ± 1.08 & 14.5 \\
Decoupled & D2NWG~\cite{D2NWG} & 60.42 ± 0.75  & \textbf{51.50 ± 0.25} & 82.42 ± 0.04 & 27.70 ± 3.24 & 32.17 ± 6.30 & 6.7 \\
Decoupled & \ourmethodo(ours) & 61.28 ± 0.32 & 48.42 ± 0.36 & 79.63 ± 0.38 & 28.90 ± 1.22 & 30.71 ± 1.29 & 4.3 \\
\rowcolor{myblue} Decoupled & \ourmethod(ours) & \textbf{64.25 ± 0.28}~~$\uparrow$2.50 & 50.85 ± 0.49~~$\downarrow$0.65 & \textbf{84.66 ± 0.26}~~$\uparrow$2.24 & \textbf{30.08 ± 1.02}~~$\uparrow$2.38 & \textbf{35.75 ± 1.18}~~$\uparrow$3.58 & \textbf{2.2}~~$\downarrow\times$3.0 \\
\hline
\end{tabular}}
\label{tab:zero_shot_performance}
\end{table*}

\begin{table*}[t]
\centering
\caption{Few-shot task accuracy comparison on Omniglot, Mini-ImageNet, and Tiered-ImageNet datasets with average per-sample evaluation latency.}
\vspace{1.5em}
\resizebox{\linewidth}{!}{
\begin{tabular}{lllllllll}
    \hline
     & & \multicolumn{2}{c}{\textbf{Omniglot}} & \multicolumn{2}{c}{\textbf{Mini-ImageNet}} & \multicolumn{2}{c}{\textbf{Tiered-ImageNet}} & \\
    \cmidrule(lr){3-4}  
    \cmidrule(lr){5-6}
    \cmidrule(lr){7-8}
    \textbf{Learning Type} & \textbf{Method} 
    & \textbf{(5, 1)} & \textbf{(5, 5)} 
    & \textbf{(5, 1)} & \textbf{(5, 5)} 
    & \textbf{(5, 1)} & \textbf{(5, 5)} 
    & \textbf{Latency (ms)} \\
    \hline
    Gradient-Based & MAML~\cite{MAML} & \textbf{98.70 ± 0.40}& \textbf{99.90 ± 0.10}& 48.70 ± 1.84 & 63.11 ± 0.92 & 48.95 ± 0.89 & 62.71 ± 0.77 & 20.9 \\
    Gradient-Based & Meta-Baseline~\cite{meta_baseline} & 97.75 ± 0.25 & 99.68 ± 0.18 & 58.10 ± 0.31 & 74.50 ± 0.29 & 68.62 ± 0.29&  83.29 ± 0.51 & 19.4 \\
    End-to-End & Meta-Hypernetwork~\cite{Meta-Hypernetwork} & 96.57 ± 0.22 & 98.83 ± 0.16 & 52.50 ± 0.28 & 67.76 ± 0.34 & 53.80 ± 0.35 & 69.98 ± 0.42 & 13.1 \\
    End-to-End & GHN3~\cite{GHN3}       & 95.23 ± 0.23 & 98.65 ± 0.19 & 63.22 ± 0.29 & 76.79 ± 0.33 & 64.72 ± 0.36 & 78.40 ± 0.46 & 17.5 \\
    Decoupled & OCD~\cite{OCD}         & 95.04 ± 0.18 & 98.74 ± 0.14 & 59.76 ± 0.27 & 75.16 ± 0.35 & 60.01 ± 0.38 & 76.33 ± 0.47 & 8.4 \\
    Decoupled & Meta-Diff~\cite{MetaDiff}    & 94.65 ± 0.65 & 97.91 ± 0.53 & 55.06 ± 0.81 & 73.18 ± 0.64 & 57.77 ± 0.90 & 75.46 ± 0.69 & 8.9 \\
    Decoupled & D2NWG~\cite{D2NWG}    & 96.77 ± 6.13 & 98.94 ± 7.49 & 61.13 ± 8.50 & 76.94 ± 6.04 & 65.33 ± 6.50 & 85.05 ± 8.25 & 10.4 \\
    Decoupled & \ourmethodo (ours) & 96.65 ± 0.19 & 99.34 ± 0.23 & 62.53 ± 0.37 & 76.25 ± 0.28 & 64.72 ± 0.17 & 83.26 ± 0.49 & 6.7 \\
    \rowcolor{myblue}Decoupled & \ourmethod (ours) & 98.25 ± 0.19~~$\downarrow$0.45 & 99.67 ± 0.13~~$\downarrow$0.23 & \textbf{65.21 ± 0.36}~~$\uparrow$1.99 & \textbf{80.17 ± 0.16}~~$\uparrow$3.23 & \textbf{69.88 ± 0.21}~~$\uparrow$1.26 & \textbf{88.36 ± 0.26}~~$\uparrow$3.31 & \textbf{3.6}~~$\downarrow\times$2.3 \\
    \hline
\end{tabular}}
\label{tab:few_shot_performance_latency}
\end{table*}

\subsubsection{Transfer Learning}
\noindent\textbf{Task.} In this task, we train and evaluate models on disjoint pre-training and evaluation datasets. During evaluation, we do not use any labeled data to adjust the models. We evaluated the transfer learning capability of the models using both accuracy and average per-sample latency.
\vspace{6pt}
\par
\noindent\textbf{Dataset.} We partitioned ImageNet-1k~\cite{imagenet} into 20k subsets of 50 classes each with 50 images per class per task for pre-training. The evaluation datasets are CIFAR-10, CIFAR-100~\cite{CIFAR1O_100}, STL-10~\cite{STL10}, Aircraft~\cite{aifcraft}, and Pets~\cite{pets}.
\vspace{6pt}
\par
\noindent\textbf{Baselines and Setups.}
We categorize the baselines into three types: gradient-based methods using conventional optimizers, end-to-end weight generation methods, and decoupled weight generation methods. We compare \ourmethod with representative baselines, including the gradient-based ICIS~\cite{ICIS}, the end-to-end GHN3~\cite{GHN3}, and the decoupled method D2NWG~\cite{D2NWG}. For all the aforementioned models, the downstream network uses ResNet12~\cite{resnet} with a linear probe for classification. For the generative model $f^G_{\phi}$, \ourmethod employs the same U-Net architecture given by Meta-Diff~\cite{MetaDiff}. In the weight preparation stage, the real optimizers SGD and Adam use a fixed learning rate of 0.005 and an automatic early-stopping strategy~\cite{early-stop} to determine the downstream task training epoch $T$. In the policy learning stage, we set the learning rate $\alpha$ and training epochs to 0.001 and 6000, respectively. The acceleration coefficient $k$ is set to 2, and the inference step $N$ is computed as $N=T/k$ for each task. \textbf{We maintain this setup across all experiments in this paper.}
\vspace{6pt}
\par
\noindent\textbf{Results.}
Table~\ref{tab:zero_shot_performance} shows that \ourmethod achieves the highest accuracy on five out of six tasks while also reducing average latency. Compared to the second-best method on each task, it yields an average accuracy improvement of 2.68\%. On CIFAR-100, its accuracy is only 0.65\% lower than the best-performing method. Compared to the fastest existing method, D2NWG, \ourmethod reduces inference latency by 3.0$\times$. As discussed in Section~\ref{sec:ablation_main}, \ourmethod also outperforms its simplified variant \ourmethodo, demonstrating the effectiveness of hybrid-policy sub-trajectory balance.

\subsubsection{Few-shot Learning}
\noindent\textbf{Task.} Following the setup provided by MAML~\cite{MAML}, we train and evaluate models on disjoint meta-training and meta-testing tasks. During the evaluation stage, we fine-tune the models on the support set of each meta-test task and measure the per-sample fine-tuning latency. We then evaluate the accuracy on the corresponding query set.
\vspace{6pt}
\par
\noindent\textbf{Dataset.} We use Omniglot~\cite{omniglot}, Mini-ImageNet~\cite{miniImagenet}, and Tiered-ImageNet~\cite{Tiered-ImageNet} datasets for the construction of 5-way 1-shot, 5-way 5-shot tasks. To evaluate generalization capabilities, we maintain distinct and separate class sets for training and evaluating phases.
\vspace{6pt}
\par
\noindent\textbf{Baselines.}\label{sec:few-shot}
Our benchmarks include gradient-based methods MAML~\cite{MAML} and Meta-Baseline~\cite{meta_baseline}, end-to-end weight generation approaches Meta-Hypernetwork~\cite{Meta-Hypernetwork}, GHN3~\cite{GHN3}, and decoupled frameworks OCD~\cite{OCD}, Meta-Diff~\cite{MetaDiff}, D2NWG~\cite{D2NWG}. Following the setting given by MAML~\cite{MAML}, the downstream neural network uses four convolution blocks with a linear probe for classification.
\vspace{6pt}
\par
\noindent\textbf{Results.}
Table~\ref{tab:zero_shot_performance} shows that \ourmethod can improve performance on almost all tasks. Since the 5-way task of Omniglot is relatively easy to learn, the gradient-based method MAML algorithm can achieve slightly higher accuracy compared to gradient-free methods. On the winning tasks, compared to the second-best method, \ourmethod achieves an average improvement of 2.45\% in accuracy. Compared to the current fastest weight generation algorithm, \ie, OCD, \ourmethod achieves a 2.3$\times$ reduction in inference latency. Note that the gradient-based methods, MAML and Meta-Baseline, exhibit high latency here due to the need for gradient computation during fine-tuning.

\subsubsection{Multi-Domain Generalization}\label{sec:multi-domain}
\noindent\textbf{Task.} In this task, we explore the domain generalization ability of our method. We follow the few-shot task setting given by H-Meta~\cite{hierar_meta} to evaluate the model's performance.
\vspace{6pt}
\par
\noindent\textbf{Dataset.} We use DomainNet~\cite{DomainNet} for the construction of 5-way 1-shot and 20-way 5-shot tasks. Specifically, we use Clipart, Infograph, Painting, Quickdraw, and Real domains for training, while Sketch domains are used for evaluation. Under this setting, the tasks in the training set may come from different domains, and the tasks in the testing set come from another unseen domain.
\vspace{6pt}
\par
\noindent\textbf{Baselines.}
We benchmark against MAML, Meta-baseline, Meta-Hypernetwork, GHN3, OCD, Meta-Diff, and D2NWG. The downstream network uses ResNet12 with a linear probe for classification.
\vspace{0.1pt}
\par
\noindent\textbf{Results.}
Table~\ref{tab:multi_domain_performance} shows that \ourmethod significantly outperforms current methods on few-shot domain generalization tasks. Compared to the second-best baselines in each task, \ourmethod achieved an average improvement of 4.64\% in accuracy. It can be observed that, compared to gradient-based methods MAML and Meta-Baseline, gradient-free methods (\ie, end-to-end and decoupled methods) exhibit a significant advantage gap. This is attributed to the generalization capability brought by the indirect weight generation method. Extending these approaches, \ourmethod incorporates local optimization policy learning, which leads to further performance improvements. \ourmethod's generalization capability stems from our learning objective, local optimization policies, which remain invariant across different datasets. In terms of overhead, \ourmethod achieves a 2.3$\times$ reduction in latency compared to OCD, showing the same advantage as in transfer learning and few-shot learning tasks.

\begin{table}[t]
\centering
\caption{Multi-domain generalization accuracy comparison on DomainNet with 5-way 1-shot and 20-way 5-shot tasks.}
\vspace{1.5em}
\resizebox{\linewidth}{!}{
\begin{tabular}{lllll}
    \toprule
          && \multicolumn{2}{c}{\textbf{DomainNet}} \\
    \cmidrule(lr){3-4}
    \textbf{Learning Type}&\textbf{Method} & \textbf{(5, 1)} & \textbf{(20, 5)}& \textbf{Latency(ms)} \\
    \midrule
    Gradient-Based&MAML~\cite{MAML}      & 45.92 ± 0.39 & 50.18 ± 0.51 & 26.8 \\
    Gradient-Based&Meta-Baseline~\cite{meta_baseline}         & 50.54 ± 0.47 & 54.45 ± 0.40 &26.2\\
    End-to-End&Meta-Hypernetwork~\cite{Meta-Hypernetwork}    & 59.00 ± 0.39 & 63.32 ± 0.35 &10.3 \\
    End-to-End&GHN3~\cite{GHN3}             & 63.11 ± 0.36 & 66.42 ± 0.33 &30.1 \\
    Decoupled&OCD~\cite{OCD} & 64.58 ± 0.42 & 67.10 ± 0.38 &9.7  \\
    Decoupled&Meta-Diff~\cite{MetaDiff}              & 64.24 ± 0.41 & 67.58 ± 0.39 &10.9 \\
    Decoupled&D2NWG~\cite{D2NWG}              & 63.68 ± 3.38 & 65.72 ± 2.85 &10.9 \\
    Decoupled&\ourmethodo(ours)        & 65.96 ± 0.47 & 67.13 ± 0.40 &7.5\\
    \rowcolor{myblue}Decoupled&\ourmethod(ours)    & \textbf{70.29 ± 0.45}~~$\uparrow$5.71 & \textbf{72.98 ± 5.40}~~$\uparrow$3.58 &\textbf{4.0}~~$\downarrow\times$2.3 \\
    \bottomrule
\end{tabular}}
\label{tab:multi_domain_performance}
\end{table}

\subsubsection{Large Language Model fine-tuning}
\noindent\textbf{Task.}
We demonstrate that \ourmethod can be applied to the fine-tuning of Large Language Models by learning to generate LoRA~\cite{LoRa} matrices for new tasks. We compared the algorithms in terms of their fine-tuning accuracy upon convergence and the latency required to achieve it.
\vspace{6pt}
\par
\noindent\textbf{Datasets.}
We conduct a case study to demonstrate the generalizability and efficiency of \ourmethod. We use five binary classification tasks, \ie, SST-2, QQP, RTE, WNIL, and CoLA from the GLUE~\cite{GLUE} benchmark for pre-training. Then we use the other two tasks, \ie, MRPC and QNIL, to evaluate the performance of the methods.
\vspace{6pt}
\par
\noindent\textbf{Baselines.}
We benchmark against Full-fine-tuning baseline, LoRA~\cite{LoRa}, AdaLoRA~\cite{AdaLoRA}, DyLoRA~\cite{DyLoRA}, and FourierFT~\cite{FourierFT}, which are all gradient-based fine-tuning algorithms. The large language model we fine-tuned is RoBERTa-base~\cite{FourierFT} and the LoRA matrices are generated following the fine-tuning process given by FourierFT~\cite{FourierFT}. The experimental results in Section~\ref{sec:multi-domain} suggest that the \ourmethod exhibits strong generalization capability, enabling it to learn across all training tasks. In contrast, gradient-based methods are single-task fine-tuning approaches that operate on one task at a time. Note that \ourmethod requires additional time to pre-train the generative model $f^G_{\phi}$; however, this is a one-time cost and demonstrates better potential in multi-task fine-tuning scenarios.
\vspace{6pt}
\par
\noindent\textbf{Results.}
Table~\ref{tab:performance_fine-tune_time_comparison} shows that \ourmethod achieves comparable binary classification accuracy on two evaluation tasks compared to other gradient-based fine-tuning algorithms while significantly accelerating the fine-tuning speed by $\times$5.7 to $\times$5.9. By implementing a decoupled framework with multiple optimizers, \ourmethod demonstrates remarkable efficiency in capturing shared local optimization policies, enabling gradient-free generation of LoRA matrices.

\section{Related Work}
\subsection{Weight Generation.}
The task of weight generation aims to generate neural network weights without gradient-based updates directly. Early approaches such as Meta-HyperNetwork~\cite{Meta-Hypernetwork}, GHN2~\cite{GHN2}, and GHN3~\cite{GHN3} adopt an end-to-end framework, where a meta-network is trained to generate weights optimized for downstream task performance. These methods, though efficient, are constrained by over-coupling between the weight generation and the task-specific objectives. This reduces the flexibility of the learned weight generator and limits the inference process to short horizons due to constraints on differentiability and tractable computation. Recent developments extend weight generation using generative models such as OCD~\cite{OCD}, Meta-Diff~\cite{MetaDiff}, and D2NWG~\cite{D2NWG}, which model optimal downstream task weights (rather than downstream task performance) via diffusion algorithms. These approaches support a long-horizon inference by treating the weight optimization process as a sampling process.

Despite their innovations, these methods focus solely on global optimal weights while neglecting the rich dynamics of intermediate optimization steps. As a result, they fail to model the local optimization policy—\ie, how weights update over time during optimization. 
\subsection{Generative Model}
Diffusion-based generative models, such as DDPM~\cite{DDPM}, Score-based Generative Models~\cite{score-based}, and Latent Diffusion~\cite{latent_diffusion}, generate data through iterative denoising and have been widely applied to continuous or structured data domains. Their forward-backward refinement processes enable strong sample quality and controllability. Their strength lies in the multi-step refinement mechanism, which offers controllability and robustness in modeling complex, structured distributions.
Bridging the conceptual gap between diffusion and discrete generative models, Zhang et al. \citep{Unifying_generative} shows that GFlowNets can be viewed as a generalized form of diffusion models, where sampling follows learned stochastic policies over compositional trajectories rather than fixed noise schedules. 
In this context, GFlowNets~\cite{GFlowNet} generate discrete trajectories leading to final states, sampling them with probabilities proportional to a reward function. To ensure consistent distributions over trajectories, the Trajectory Balance (TB) and sub-Trajectory Balance (sub-TB) objectives~\cite{TB} were proposed. TB and sub-TB ensure equality between the forward and backward trajectory probabilities, scaled by a learnable flow term. 

Despite these advances, diffusion models and GFlowNet-like methods typically rely on on-policy learning: models are supervised only by self-sampled trajectories and final reward, limiting their capacity to align generated intermediate states with off-policy data or external supervision signals.

\begin{table}[t]
\centering
\caption{Accuracy and fine-tuning latency comparison on GLUE-MRPC and GLUE-QNLI tasks with different fine-tuning algorithms.}
\vspace{1em}
\resizebox{\linewidth}{!}{
\begin{tabular}{llllll}
    \toprule
     && \multicolumn{2}{c}{\textbf{MRPC}} & \multicolumn{2}{c}{\textbf{QNLI}} \\
     \cmidrule(lr){3-4}
    \cmidrule(lr){5-6}
    \textbf{fine-tuning Type}&\textbf{Method} & \textbf{Acc} & \textbf{Latency (h)} & \textbf{Acc} & \textbf{Latency (h)} \\
    \midrule
    Gradient-Based&Full-fine-tune   & \textbf{90.24 ± 0.57} & 1.47 & 92.84 ± 0.26 & 3.15 \\
    Gradient-Based&LoRA~\cite{LoRa}            & 89.76 ± 0.69 & 0.81 & \textbf{93.32 ± 0.20} & 1.76 \\
    Gradient-Based&AdaLoRA~\cite{AdaLoRA}         & 88.71 ± 0.73 & 0.74 & 93.17 ± 0.25 & 1.68 \\
    Gradient-Based&DyLoRA~\cite{DyLoRA}          & 89.59 ± 0.81 & 0.76 & 92.21 ± 0.32 & 1.63 \\
    Gradient-Based&FourierFT~\cite{FourierFT}       & 90.03 ± 0.54 & 0.68 & 92.25 ± 0.15 & 1.55 \\
    Gradient-Free&\ourmethodo(ours)        & 87.59 ± 0.57 & 0.25 & 90.48 ± 0.27 & 0.51 \\
    \rowcolor{myblue}Gradient-Free&\ourmethod(ours)       & 89.62 ± 0.66~~$\downarrow$0.62 & \textbf{0.12}~~$\downarrow\times$5.7 & 92.38 ± 0.29~~$\downarrow$0.94 & \textbf{0.26}~~$\downarrow\times$5.9 \\
    \bottomrule
\end{tabular}}
\label{tab:performance_fine-tune_time_comparison}
\end{table}

\section{Conclusion}
In this work, we target the problem of weight generation by viewing it as an optimization policy learning problem. We propose \ourmethod, a novel weight generation framework for the issues of over-coupling and long horizon. To address the limited flexibility caused by over-coupling, our method adopts a decoupled two-stage learning process that enables the learning of diverse local optimization policies. To address the inefficiency in inference caused by the long-horizon issue. \ourmethod introduces Hybrid-Policy Sub-Trajectory Balance to capture local optimization policies. Theoretically, we demonstrate that focusing solely on local optimization policy learning addresses the long-horizon issue while also enhancing the generation of global optimal weights. Empirically, we demonstrate \ourmethod’s superior accuracy and inference efficiency in tasks that require frequent weight updates, such as transfer learning, few-shot learning, domain adaptation, and large language model adaptation.

\section*{Acknowledgements}
This work was supported by the China Scholarship Council (CSC) under Grant No.~202406160071, the Pioneer Centre for AI, DNRF grant number P1, the National Key Research and Development Program of China under Grant No.~2023YFB4502701, and the National Natural Science Foundation of China under Grant No.~62232007.

\bibliography{ecai-sample-and-instructions}

\end{document}